%% file: acl_latex.tex
\pdfoutput=1

\documentclass
[11pt]{article}

\usepackage[final]{acl}

\usepackage{times}
\usepackage{latexsym}

\usepackage[T1]{fontenc}

\usepackage[utf8]{inputenc}

\usepackage{microtype}

\usepackage{inconsolata}

\usepackage{graphicx}
\usepackage{amssymb}
\usepackage{bm}
\usepackage{amsmath}
\usepackage{algorithm}
\usepackage{algpseudocode}
\usepackage{booktabs}
\usepackage{multirow}
\usepackage{graphicx}
\usepackage{bbding}
\usepackage{enumitem}
\setlist[itemize]{nosep,leftmargin=*}

\title{DALK: \underline{D}ynamic Co-\underline{A}ugmentation of \underline{L}LMs and \underline{K}G to answer Alzheimer’s Disease Questions with Scientific Literature}

\author{\textbf{Dawei Li\textsuperscript{1}\thanks{\ \ Equal Constributions}, Shu Yang\textsuperscript{2}$^{*}$, Zhen Tan\textsuperscript{1}, Jae Young Baik\textsuperscript{2},} \\ 
\textbf{Sukwon Yun\textsuperscript{3}, Joseph Lee\textsuperscript{2}, Aaron Chacko\textsuperscript{2}, Bojian Hou\textsuperscript{2},}\\
\textbf{Duy Duong-Tran\textsuperscript{2,4}, Ying Ding\textsuperscript{5}, Huan Liu\textsuperscript{1}\thanks{\ \ Corresponding authors}, Li Shen\textsuperscript{2\dag}, Tianlong Chen\textsuperscript{3\dag}} \\
  \textsuperscript{1}School of Computing, and Augmented Intelligence, Arizona State University \\
  \textsuperscript{2}University of Pennsylvania Perelman School of Medicine \\
  \textsuperscript{3}Department of Computer Science, The University of North Carolina at Chapel Hill \\
  \textsuperscript{4}Department of Mathematics, United States Naval Academy \\
  \textsuperscript{5}School of Information, The University of Texas at Austin, Austin \\
}

\begin{document}
\maketitle
\begin{abstract}
\input{latex/0-abstract}
\end{abstract}

\section{Introduction}
\input{latex/1-introduction_v2}

\section{Related Work}
\input{latex/2-related-work_v2}

\section{Our Methodology}
\input{latex/3-our-methodology}

\section{Main Experiment}
\input{latex/4-main-experiment}

\section{Further Analysis}
\input{latex/5-further-analysis}

\section{Conclusion}
\input{latex/6-conclusion}

\section{Limitations and Future Work}
\input{latex/7-limitations}

\section{Ethics Statement}
\input{latex/8-ethics-statement}

\section*{Acknowledgments}
This work was supported in part by the NIH grants U01 AG066833, U01 AG068057, R01 AG071470, and P30 AG073105.
The material in this presentation is also based upon work supported by the U.S. Department of Homeland Security under Grant Award Number, 17STQAC00001-08-00 and the U.S. Office of Naval Research (ONR) under grant N00014-21-1-4002.
\bibliography{custom}

\appendix
\onecolumn

\section{AD-KG Evaluation}

We conduct a manual evaluation of our AD-KG to guarantee its quality. We randomly sample 100 triples from AD-KG constructed with generative and RE methods and ask the annotator to check whether the fact in each triple is valid. The annotator is encouraged to find evidence from the original abstract corpus or by searching the web. The results are presented in Table~\ref{manual evaluation}. We found that while both AD-KGs achieve acceptable accuracy, the RE construction method produces an AD-KG of higher quality. This further validates our conclusion in Section~\ref{Ablation Study on KG Construction} regarding the trade-off between coverage and accuracy.

\begin{table}[h]
\centering
\begin{tabular}{lc}
\hline
                                 & Accuracy \\ \hline
AD-KG w/ Generative Construction & 91\%     \\
AD-KG w/ RE Construction         & 83\%    \\ \hline
\end{tabular}
\caption{Manual evaluation result on AD-KG.}
\label{manual evaluation}
\end{table}

\section{Details of LLMs for KG}
\label{Details of LLMs for KG}
Table~\ref{example:generative} and~\ref{example:RE} present examples of our two KG construction methods respectively.
For both methods, we adopt a select-or-generate prompt to instruct the LLM whether to choose a relation from hetionet~\cite{himmelstein2017systematic}, a well-built general medical KG, or generate a new one to describe the relationship between two entities.
In the RE construction method, we also conduct a type matching (Table~\ref{match:type}) for each entity from type name of PubTator to that of Hetionet and ask the LLM to choose from the relation set that corresponds to the two entities' types (Table~\ref{match:relation}).

\begin{table*}[h]
\centering
\begin{tabular}{lp{5in}p{5in}}
\hline
Input  & Read the following abstract, extract the relationships between each entity.You can choose the relation from: (covaries, interacts, regulates, resembles, downregulates, upregulates, associates, binds, treats, palliates), or generate a new predicate to describe the relationship between the two entities. Output all the extract triples in the format of "head | relation | tail". For example: "Alzheimer's disease | associates | memory deficits" Abstract: Thiamine pyrophosphate (TPP) and the activities of thiamine-dependent enzymes are reduced in Alzheimer's disease (AD) patients. In this study, we analyzed the relationship between thiamine deficiency (TD) and amyloid precursor protein (APP) processing in both cellular and animal models of TD. In SH-SY5Y neuroblastoma cells overexpressing APP, TD promoted maturation of beta-site APP cleaving enzyme 1 (BACE1) and increased beta-secretase activity which resulted in elevated levels of beta-amyloid (Abeta) as well as beta-secretase cleaved C-terminal fragment (beta-CTF). An inhibitor of beta-secretase efficiently reduced TD-induced up-regulation of Abeta and beta-CTF. Importantly, thiamine supplementation reversed the TD-induced alterations. Furthermore, TD treatment caused a significant accumulation of reactive oxygen species (ROS); antioxidants suppressed ROS production and maturation of BACE1, as well as TD-induced Abeta accumulation. On the other hand, exogenous Abeta(1-40) enhanced TD-induced production of ROS. A study on mice indicated that TD also caused Abeta accumulation in the brain, which was reversed by thiamine supplementation. Taken together, our study suggests that TD could enhance Abeta generation by promoting beta-secretase activity, and the accumulation of Abeta subsequently exacerbated TD-induced oxidative stress.Entity: Thiamine (thiamine), deficiency increases beta-secretase, TPP (Thiamine pyrophosphate), Alzheimer's disease (AD), patients, amyloid precursor protein (beta-amyloid (Abeta)), neuroblastoma, BACE1 (beta-site APP cleaving enzyme 1), Abeta, ROS (reactive oxygen species), Abeta accumulation, mice. Output: \\ \hline
Output & Thiamine deficiency | downregulates | TPP, Thiamine deficiency | regulates | Alzheimer's disease, Alzheimer's disease | involves | beta-secretase, Amyloid precursor protein | produces | beta-amyloid, Amyloid precursor protein | in | neuroblastoma, BACE1 | is | beta-secretase, BACE1 | regulates | beta-amyloid, Abeta | is | beta-amyloid, Abeta | in | mice, Abeta | causes | reactive oxygen species, Abeta | regulates | ROS \\ \hline                                                   \end{tabular}
\caption{An example to extract knowledge triples using generative construction method.}
\label{example:generative}
\end{table*}

\begin{table*}[h]
\centering
\begin{tabular}{lp{5in}p{5in}}
\hline
Input  & Read the following abstract, answer the following question. Abstract: Thiamine pyrophosphate (TPP) and the activities of thiamine-dependent enzymes are reduced in Alzheimer's disease (AD) patients. In this study, we analyzed the relationship between thiamine deficiency (TD) and amyloid precursor protein (APP) processing in both cellular and animal models of TD. In SH-SY5Y neuroblastoma cells overexpressing APP, TD promoted maturation of beta-site APP cleaving enzyme 1 (BACE1) and increased beta-secretase activity which resulted in elevated levels of beta-amyloid (Abeta) as well as beta-secretase cleaved C-terminal fragment (beta-CTF). An inhibitor of beta-secretase efficiently reduced TD-induced up-regulation of Abeta and beta-CTF. Importantly, thiamine supplementation reversed the TD-induced alterations. Furthermore, TD treatment caused a significant accumulation of reactive oxygen species (ROS); antioxidants suppressed ROS production and maturation of BACE1, as well as TD-induced Abeta accumulation. On the other hand, exogenous Abeta(1-40) enhanced TD-induced production of ROS. A study on mice indicated that TD also caused Abeta accumulation in the brain, which was reversed by thiamine supplementation. Taken together, our study suggests that TD could enhance Abeta generation by promoting beta-secretase activity, and the accumulation of Abeta subsequently exacerbated TD-induced oxidative stress.Entity: Thiamine (thiamine), deficiency increases beta-secretase, TPP (Thiamine pyrophosphate), Alzheimer's disease (AD), patients, amyloid precursor protein (beta-amyloid (Abeta)), neuroblastoma, BACE1 (beta-site APP cleaving enzyme 1), Abeta, ROS (reactive oxygen species), Abeta accumulation, mice. Question: predict the relationship between Disease entity "sclerosis" and Disease entity "multiple sclerosis", first choose from the following options: A. resembles B. no-relation C. others, please specify by generating a short predicate in 5 words. Answer: Let's think step by step: \\ \hline
Output & 1. Sclerosis is a disease of the central nervous system. 2. Multiple sclerosis is the most common form of sclerosis. 3. So the relationship between sclerosis and multiple sclerosis should be "sclerosis is a kind of multiple sclerosis". So the answer is: A. resembles. So the answer is: A. resembles  \\ \hline         
\end{tabular}
\caption{An example to extract the relationship between ``sclerosis'' and ``multiple sclerosis'' using RE construction method.}
\label{example:RE}
\end{table*}

\begin{table*}[h!]
\centering
\begin{tabular}{ll}
\hline
PubTator Type       & Hetionet Type        \\ \hline
Gene              & genes                \\
Chemical          & compounds            \\
Disease           & diseases             \\ \hline
\end{tabular}
\caption{Entity type match from PubTator to Hetionet.}
\label{match:type}
\end{table*}

\begin{table*}[h!]
\centering
\begin{tabular}{ll}
\hline
Type-Type           & Relations                                        \\ \hline
genes-genes         & {[}"covaries", "interacts", "regulates"{]}       \\
disease-disease     & {[}"resembles"{]}                                \\
compounds-compounds & {[}"resembles"{]}                                \\
genes-diseases      & {[}"downregulates","associates","upregulates"{]} \\
genes-compounds     & {[}"binds", "upregulates", "downregulates"{]}    \\
compounds-diseases  & {[}"treats", "palliates"{]}                      \\ \hline
\end{tabular}
\caption{Type-type to relation match in Hetionet.}
\label{match:relation}
\end{table*}

\newpage
\section{Details of KG for LLMs}
\label{Details of KG for LLMs}

In this section, we provide detailed input and output for adopting KG to augment LLMs, including path-based and neighbor-based sub-graph sampling results (Table~\ref{tab:sub-graph}), self-aware knowledge retrieval (Table~\ref{tab:detailed self}), describing sub-graphs with LLMs (Table~\ref{tab:self-describe}) and inference with sampled knowledge (Table~\ref{tab:inference}).
The question we showcase here is ``The area of the brain resistant to Neurofibrillary tangles of Alzheimer’s disease is: A. Visual association areas B. Entorhinal coex C. Temporal lobe D.Lateral geniculate body'', which is the same as the one we use in Section~\ref{Case Study}.

\begin{table*}[!h]\small
\centering
\begin{tabular}{ll}
\hline
Path-based Sub-graph     & \begin{tabular}[c]{@{}l@{}}neurofibrillary tangles-\textgreater{}FORM BY-\textgreater{}microtubule-associated protein tau-\textgreater{}BINDS-\textgreater\\ (18)F-THK-5117-\textgreater{}ADMINISTERED TO-\textgreater{}rats-\textgreater{}has-\textgreater{}Alzheimer's disease -\textgreater\\ Alzheimer's disease -\textgreater{}affects-\textgreater{}human-\textgreater{}has-\textgreater{}AD-\textgreater{}DISEASE OF-\textgreater{}Brain\\ entorhinal cortex-\textgreater{}is a part of-\textgreater{}brain-\textgreater{}ASSOCIATES-\textgreater\\ mouse with Alzheimer's disease-\textgreater{}brain region-\textgreater{}temporal lobe\end{tabular} \\ \hline
Neighbor-based Sub-graph & \begin{tabular}[c]{@{}l@{}}neurofibrillary tangles-\textgreater{}FORM BY-\textgreater{}microtubule-associated protein tau\\ Alzheimer's disease -\textgreater{}causes-\textgreater{}neuronal death\\ Alzheimer's disease -\textgreater{}associates-\textgreater{}cognitive decline\\ Alzheimer's disease -\textgreater{}affects-\textgreater{}human\\ Alzheimer's disease -\textgreater{}has subtype-\textgreater{}neurodegenerative diseases\end{tabular}           \\ \hline                                        \end{tabular}
\caption{An example of path-based and neighbor-based sub-graph for the question.}
\label{tab:sub-graph}
\end{table*}

\begin{table}[!h]\small
\centering
\begin{tabular}{ll}
\hline
Input  & \begin{tabular}[c]{@{}l@{}}There is a question and some knowledge graph. The knowledge graphs follow entity-\textgreater{}relationship\\ -\textgreater{}entity list format.\\ Graph: \\ neurofibrillary tangles-\textgreater{}FORM BY-\textgreater{}microtubule-associated protein tau-\textgreater{}BINDS-\textgreater\\ (18)F-THK-5117-\textgreater{}ADMINISTERED TO-\textgreater{}rats-\textgreater{}has-\textgreater{}Alzheimer's disease -\textgreater{}Alzheimer's disease -\textgreater\\ affects-\textgreater{}human-\textgreater{}has-\textgreater{}AD-\textgreater{}DISEASE OF-\textgreater{}Brain\\ entorhinal cortex-\textgreater{}is a part of-\textgreater{}brain-\textgreater{}ASSOCIATES-\textgreater\\ mouse with Alzheimer's disease-\textgreater{}brain region-\textgreater{}temporal lobe\}\\ \\ Question: \\ \\ The area of the brain resistant to Neurofibrillary tangles of Alzheimer’s disease is: \\ A. Visual association areas \\ B. Entorhinal coex \\ C. Temporal lobe \\ D.Lateral geniculate body\\ \\ \\ Please rerank the knowledge graph and output at most 5 important and relevant triples for\\ solving the given question. Output the reranked knowledge in the following format:\\ Reranked Triple1: xxx ——\textgreater xxx\\ Reranked Triple2: xxx ——\textgreater xxx\\ Reranked Triple3: xxx ——\textgreater xxx\\ Reranked Triple4: xxx ——\textgreater xxx\\ Reranked Triple5: xxx ——\textgreater xxx\\ \\ Answer:\end{tabular} \\ \\ \hline
Output & \begin{tabular}[c]{@{}l@{}}Reranked Triples1: entorhinal cortex -\textgreater{}is a part of -\textgreater{}brain\\ Reranked Triples2: entorhinal cortex -\textgreater{}associates -\textgreater{}mouse with Alzheimer’s disease\\ Reranked Triples3: temporal lobe -\textgreater{}affected by -\textgreater{}Alzheimer’s disease\end{tabular}    \\ \hline                      \end{tabular}
\caption{An example of our self-aware knowledge retrieval in path-based sub-graph.}
\label{tab:detailed self}
\end{table}


\begin{table}[!h]\small
\centering
\begin{tabular}{ll}
\hline
Input  & \begin{tabular}[c]{@{}l@{}}There are some knowledge graph paths. They follow entity-\textgreater{}relationship-\textgreater{}entity format.\\ \\ Reranked Triples1: entorhinal cortex -\textgreater{}is a part of -\textgreater{}brain\\ Reranked Triples2: entorhinal cortex -\textgreater{}associates -\textgreater{}mouse with Alzheimer’s disease\\ Reranked Triples3: temporal lobe -\textgreater{}affected by -\textgreater{}Alzheimer’s disease\\ \\ Use the knowledge graph information. Try to convert them to natural language, respectively. \\ Use single quotation marks for entity name and relation name. \\ And name them as Path-based Evidence 1, Path-based Evidence 2,...\\ \\ Output:\end{tabular} \\ \hline
Output & \begin{tabular}[c]{@{}l@{}}Path-based Evidence 1: 'Entorhinal cortex' is a part of 'brain'.\\ Path-based Evidence 2: 'Entorhinal cortex' associates 'mouse' with 'Alzheimer's disease'.\\ Path-based Evidence 3: 'Temporal lobe' is affected by 'Alzheimer's disease'.\end{tabular}        \\ \hline                 \end{tabular}
\caption{An example to describe the self-retrieved path-based sub-graph with the LLM.}
\label{tab:self-describe}
\end{table}


\begin{table}[!h]\small
\centering
\begin{tabular}{ll}
\hline
Input  & \begin{tabular}[c]{@{}l@{}}Question: The area of the brain resistant to Neurofibrillary tangles of Alzheimer's disease is:\\ A.Visual association areas\\ B.Entorhinal coex\\ C.Temporal lobe\\ D.Lateral geniculate body\\ \\ You have some medical knowledge information in the following:\\ \#\#\#Path-based Evidence 1: 'Entorhinal cortex' is a part of 'brain'.\\ Path-based Evidence 2: 'Entorhinal cortex' associates 'mouse' with 'Alzheimer's disease'.\\ Path-based Evidence 3: 'Temporal lobe' is affected by 'Alzheimer's disease'.\\ \#\#\#Neighbor-based Evidence 1: 'neurofibrillary tangles' are 'FORMED BY' '\\ microtubule-associated protein tau'.\\ Neighbor-based Evidence 2: 'Alzheimer's disease' 'CAUSES' 'neuronal death'.\\ Neighbor-based Evidence 3: 'Alzheimer's disease' 'ASSOCIATES' with 'cognitive decline'.\\ Neighbor-based Evidence 4: 'Alzheimer's disease' 'AFFECTS' 'human'.\\ Neighbor-based Evidence 5: 'Alzheimer's disease' 'HAS SUBTYPE' 'neurodegenerative diseases'."\\ \\ Answer: Let's think step by step:\end{tabular} \\ \hline
Output & \begin{tabular}[c]{@{}l@{}}In Alzheimer's disease, neurofibrillary tangles are composed of \\ abnormal aggregations of tau protein within neurons, leading to \\ neuronal dysfunction and eventual cell death. The lateral geniculate \\ body is a relay center in the thalamus that processes visual \\ information from the retina to the visual cortex. Since the lateral \\ geniculate body is not directly involved in memory and cognitive \\ function, it is less likely to be affected by neurofibrillary tangles in \\ Alzheimer's disease compared to other brain regions such as the \\ entorhinal cortex and temporal lobe which play crucial roles in memory \\ and cognitive processing. Therefore, the area of the brain resistant to \\ neurofibrillary tangles of Alzheimer's disease is the lateral geniculate \\ body (option D)\end{tabular} \\ \hline
\end{tabular}
\caption{An example of inference with selected knowledge}
\label{tab:inference}
\end{table}

\newpage
\section{ADQA Benchmark}
\label{ADQA Benchmark}

Table~\ref{tab:self-sample} shows the detailed prompt of our self-sample method to construct ADQA benchmark.
Table~\ref{stat: ADQA} gives the statistics of our ADQA benchmark.

\begin{table}[!h]
\centering
\begin{tabular}{ll}
\hline
Input  & \begin{tabular}[c]{@{}l@{}}Judge whether the question below is related to Alzheimer's Disease. Please answer yes or no. \\ Question: Treatable causes of dementia are \_\_\_.\\ a).AD b).Hypothyroidism c).Multi-infarct dementia d).SDH e).Hydrocephalus\\ Is the question related to Alzheimer's Disease? Answer:\end{tabular} \\ \hline
Output & Yes \\ \hline              
\end{tabular}
\caption{An example from MedMCQA to self-sample AD-related QA sample with LLMs.}
\label{tab:self-sample}
\end{table}

\begin{table}[!h]
\centering
\begin{tabular}{lccccc}
\hline
Dataset & MedQA & MedMCQA & MMLU & QA4MRE & Total \\ \hline
Number & 152   & 210     & 49   & 35     & 446  \\ \hline
\end{tabular}
\caption{Statistics of our ADQA benchmark.}
\label{stat: ADQA}
\end{table}

\section{Further Experiment for RAG}
\label{Further Experiment for RAG}

\begin{table}[!h]
\centering
\begin{tabular}{lccccc}
\hline
& MedQA                                               & MedMCQA                                             & NMMLU                                               & QA4MRE                                              & AVG                                                                     \\ \hline
Almanac w/ 256 chunk size   & 50.0                                                & 69.0                                                & 67.3                                                & 62.9                                                & 62.3                                                                    \\
Almanac w/ top 10 document  & 48.7                                                & 68.6                                                & 65.3                                                & 62.9                                                & 61.4                                                                    \\
Almanac w/ CoT              &  50.0 & 65.7 &  77.6 &  65.7 & 64.7 \\
Clinfo.ai w/ 256 chunk size & 48.6                                                & 66.7                                                & 81.6                                                & 65.7                                                & 65.7                                                                    \\
Clinfo.ai w/ top 5 document & 43.4                                                & 68.1                                                & 77.6                                                & 68.6                                                & 64.4                                                                    \\
Clinfo.ai w/ CoT            & 48.7                        & 68.6                        & 79.6                        & 68.6                        & \multicolumn{1}{r}{65.0}      \\ \hline                 
\end{tabular}
\caption{Further experiment in RAG methods with different hyper-parameter settings.}
\label{tab:further exp with RAG}
\end{table}

\section{Time Cost for Subgraphs Extraction}
\label{Time Cost for Subgraphs Extraction}

\begin{table}[!t]
\centering
\begin{tabular}{lccccc}
\hline
& MedQA                                               & MedMCQA                                             & NMMLU                                               & QA4MRE                                              & AVG                                                                     \\ \hline
AVG Length   & 107.4                                                & 23.8                                                & 342.9                                                & 17.6                                                & 122.9                                                                    \\
Time Cost (s)  & 2.25                                                & 0.89                                                & 2.25                                                & 1.09                                                & 1.62                                                                \\ \hline                 
\end{tabular}
\caption{Average lengths of questions and time costs to construct the two subgraphs (path-based subgraph and neighbor-based subgraph) for each dataset.}
\label{tab:further exp with RAG}
\end{table}
\end{document}

%% file: latex/0-abstract.tex
Recent advancements in large language models (LLMs) have achieved promising performances across various applications.
Nonetheless, the ongoing challenge of integrating long-tail knowledge continues to impede the seamless adoption of LLMs in specialized domains.
In this work, we introduce DALK, a.k.a. \underline{D}ynamic Co-\underline{A}ugmentation of \underline{L}LMs and \underline{K}G, to address this limitation and demonstrate its ability on studying Alzheimer's Disease (AD), a specialized sub-field in biomedicine and a global health priority. With a synergized framework of LLM and KG mutually enhancing each other, we first leverage LLM to construct an evolving AD-specific knowledge graph (KG) sourced from AD-related scientific literature, and then we utilize a coarse-to-fine sampling method with a novel self-aware knowledge retrieval approach to select appropriate knowledge from the KG to augment LLM inference capabilities.
The experimental results, conducted on our constructed AD question answering (ADQA) benchmark, underscore the efficacy of DALK. 
Additionally, we perform a series of detailed analyses that can offer valuable insights and guidelines for the emerging topic of mutually enhancing KG and LLM. 
We will release the code and data at https://github.com/David-Li0406/DALK.

%% file: latex/1-introduction_v2.tex
Alzheimer’s Disease (AD) is a neurodegenerative disorder characterized by progressive declines in cognitive and functional status over a span of decades~\cite{AD}. 
However, current AD therapy developments are facing critical challenges 
due to the lack of knowledge and understanding of the underlying etiological mechanisms of the disease. 
Although scientific literature and dedicated biomedical databases could supply rich sources of AD knowledge, manual review of relevant information is impossible due to the large volume.

As large language models~(LLMs)~\citep{brown2020language,zhang2022opt,anil2023palm,touvron2023llama} with chain-of-thought (CoT)-based prompting~\cite{wei2022chain,wang2022self,tong2023eliminating,yao2023tree,besta2023graph} demonstrate strong language capabilities across various tasks, there have been attempts to leverage LLMs-based systems in general biomedical and AD-related applications~\cite{mao2023ad,li2023two, yan2024leveraging,feng2023large}.
However, while the LLMs have shown promising performances in many general tasks, recent studies revealed LLMs' limitations in long-tail~\cite{kandpal2023large} and domain-specific~\cite{li2023multi,li2024contextualization} knowledge, thereby significantly impeding their adaptations in vertical fields such as AD.
To deal with this issue, the most common strategies are retrieval augmented generation (RAG) and domain-specific LLMs training.

Nevertheless, directly applying these strategies in the context like AD would still suffer from several issues.
First, \textbf{Data Quality}: Same as many biomedical fields, scientific literature composes the largest publicly available corpus source in AD.
Yet, the dense and information-overloaded nature of scientific literature, when combined with raw text retrieval methods, can lead to the retrieval of irrelevant and noisy information. 
Previous research has shown that noisy and irrelevant corpora can significantly undermine the performance of LLMs~\cite{yu2023chain,chen2024benchmarking,wu2024easily}.
Second, \textbf{Efficiency \& Scale Issues}: Being an critical field of research, the knowledge of AD is rapidly evolving with scientific advancements at a remarkable pace and scale.
However, retraining a domain-specific LLM or updating certain knowledge in it demands substantial computational resources~\cite{hu2021lora,ovadia2023fine,zhang2024balancing}.
This efficiency issue would also limit the sizes of domain-specific LLMs, consequently affecting their performances. 

To tackle these limitations, here we propose a \underline{D}ynamic Co-\underline{A}ugmentation of \underline{L}LMs and \underline{K}G (DALK) framework that facilitates mutual benefits between LLMs and knowledge graphs (KG) for the AD domain.
Initially, our framework addresses the data quality challenge by extracting more structural and accurate knowledge from unstructured and dense scientific literature and constructing a domain-specific knowledge graph tailored to AD.
We employ two widely utilized knowledge graph construction methods, namely pair-wise construction~\cite{carta2023iterative,wadhwa2023revisiting} and generative construction~\cite{han2023pive,bi2024codekgc}, to comprehensively assess their impact on knowledge graph quality.
Then, we adopt a coarse-to-fine sampling method with a novel self-aware knowledge retrieval approach to select appropriate knowledge from the knowledge graph and thus further address the data quality problem.
Notably, the tuning-free nature of our framework significantly enhances efficiency and facilitates its application in large-scale and API-based language models~\cite{Introducing}.
In the evaluation section, we derive an Alzheimer's Disease question answering (ADQA) benchmark from existing general medical QA datasets with millions of samples filtered by a curated keyword list and self-sampling of LLMs.
Our extensive experiment on ADQA demonstrates the effectiveness of our framework in domain-specific applications compared with general biomedical LLMs and retrieval augmented models.
Further evaluation and analysis provide valuable insights into constructing high-quality knowledge graphs and sampling accurate knowledge from them.


In summary, our contribution in this work can be summarized as follows:
\begin{itemize}
    \item We identify the constraints of the current methods for LLMs in domain-specific areas like AD and introduce DALK, a co-augmentation framework of the LLM and KG to address these issues.
    \item We build AD-specific KG and QA benchmark. Through extensive comparisons with other methods, we showcase the effectiveness of DALK.
    \item We delve into a comprehensive analysis of our proposed method and provide valuable insights and guidance on how to construct a high-quality KG and sample accurate knowledge from it.
\end{itemize}

%% file: latex/2-related-work_v2.tex
\paragraph{The interplay between LLMs and KGs}
KGs~\cite{miller1995wordnet,speer2017conceptnet,vrandevcic2014wikidata,xiong2023tilp,xiong2024teilp} serve as structured representations of factual knowledge, typically expressed as (head, relation, tail) triples.
Their structured, factual, and interpretable nature renders them excellent complements to parametric language models~\cite{pan2024unifying,Zhu2023LLMsFK}.
Recently, with the rise of large language models (LLMs), numerous studies have delved into exploring the synergy between LLMs and KGs for various purposes~\cite{pan2024unifying,tan2024large}.
There are a lot of efforts in conducting knowledge graph construction~\cite{carta2023iterative,wadhwa2023revisiting,han2023pive,bi2024codekgc,datta2024construction}, completion~\cite{wei2023kicgpt,zhang2023making,li2024contextualization} with the aid of LLMs.
Conversely, other works aim to enhance LLMs by integrating knowledge sampled from KGs during both training~\cite{tang2023graphgpt,luo2024knowla,dernbach2024glam,rangel2024sparql} and inference~\cite{kim2023kg,wen2023mindmap,jiang2023structgpt,sun2023think} times.
Our work distinguishes itself by proposing a co-augmentation framework for LLMs and KGs, facilitating their mutual enhancement, and applying it to the domain of AD.

\paragraph{LLMs and KGs for AD research}
LLMs and KGs have both been applied to Alzheimer's Disease research in previous studies. 
Pre-trained language models are utilized to work on AD detection and many other related tasks based on speech recordings and transcripts~\cite{balagopalan2020to, agbavor2022predicting}, electronic health records (EHRs)~\cite{mao2023ad,li2023two, yan2024leveraging}, and tabular data~\cite{feng2023large}. KGs have been widely used in biomedical research, yet only a few are specifically for AD research~\cite{romano2023the, pu2023graph, hsieh2023synthesize, yi2022mining, Geesa2023in}. These KGs were generally constructed from a variety of information derived from heterogeneous biomedical databases (e.g. for genes, drugs, pathways, etc.) or scientific literature related to AD. Despite the aforementioned efforts for LLMs and KGs in AD research, no prior study has explored using LLM to augment AD-KG, or vice versa, let alone the potential for mutual enhancement between the two as we propose here.

%% file: latex/3-our-methodology.tex
\begin{figure*}[!t]
    \centering
    \includegraphics[width=16cm]{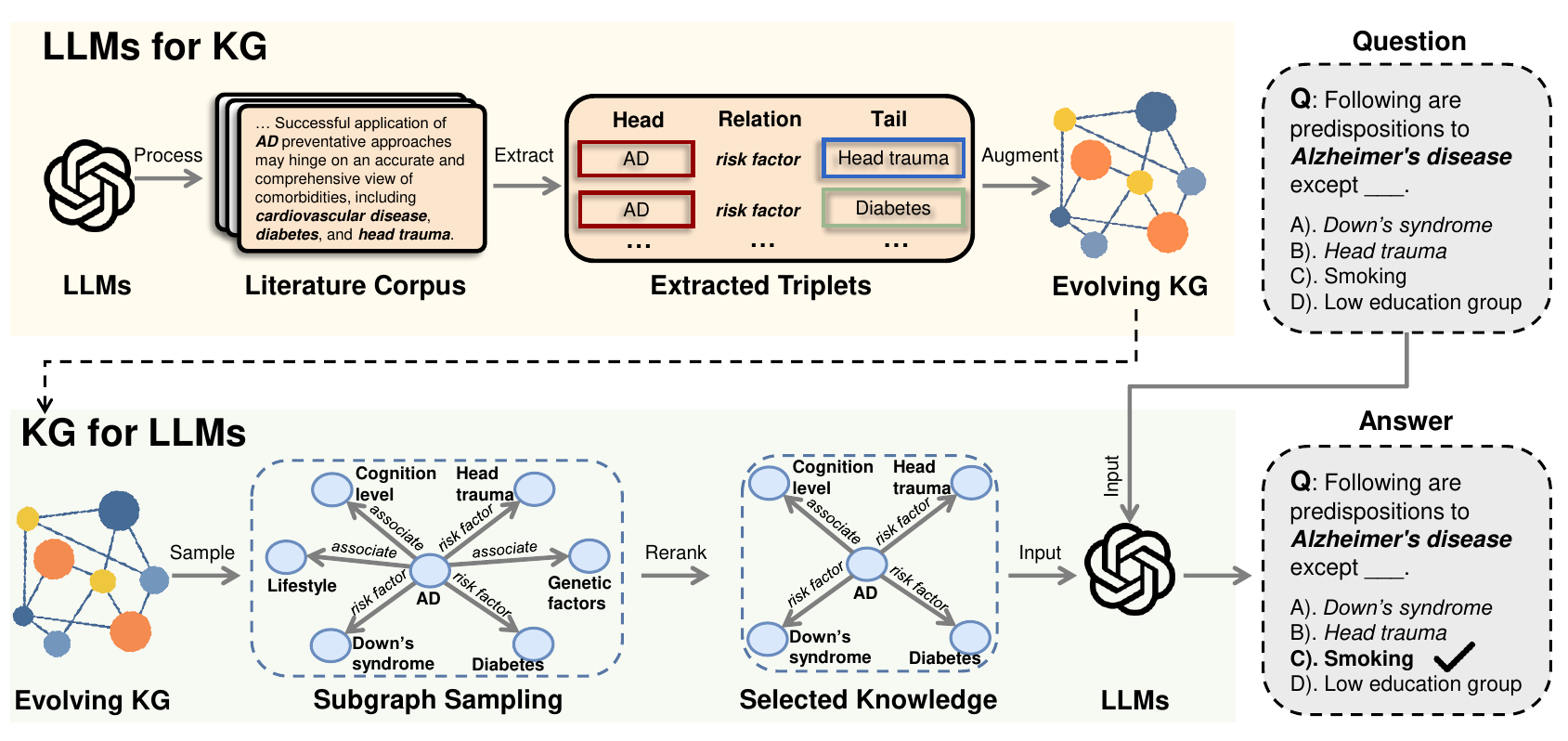}
    \caption{The overview pipeline of DALK. We first extract structural knowledge from unstructured corpora and construct a domain-specific knowledge graph tailored to AD (Section~\ref{LLMs for KG}). Then, we utilize a coarse-to-fine sampling method with a novel self-aware knowledge retrieval approach to select appropriate knowledge from the knowledge graph (Section~\ref{KG for LLMs}).}
    \label{fig:pipeline}
\end{figure*}

This section elaborates on our dynamic co-augmentation framework of LLMs and KG. Section~\ref{LLMs for KG} presents the details of augmenting an AD-specific evolving KG with LLMs and literature corpus in a time-slicing fashion (i.e. year by year). Following it, Section~\ref{KG for LLMs} describes the process of sampling appropriate knowledge from the evolving KG to enhance LLMs' reasoning. Figure~\ref{fig:pipeline} illustrates an overall pipeline of our method DALK.

\begin{table}[h!]
\centering
\begin{tabular}{lcc}
\hline
            & \multicolumn{1}{l}{$KG_{pair}$} & \multicolumn{1}{l}{$KG_{gen}$} \\
\hline
\#Corpus    & 9,764                           & 9,764                                        \\
\#Nodes     & 13,509                           & 20,545                          \\
\#Relations & 3,952                            & 3,651                           \\
\#Triples   & 171,431                          & 53,585                          \\
\hline
\end{tabular}
\caption{Detailed statistics about AD-KG.}
\label{statistics}
\end{table}

\subsection{LLMs for KG}
\label{LLMs for KG}

\paragraph{Corpus Collection} To create an AD-specific knowledge graph, we follow~\cite{pu2023graph} and use the AD corpus collected by a domain expert Professor Colin Masters at the University of Melbourne who discovered amyloid proteins being the potential cause of AD~\cite{masters1985amyloid}. The corpus is based on his extensive bibliography of representative AD-related papers and consists of more than 16K PMID (PubMed ID)-indexed articles from 1977 to 2021. For our study, we focus on the papers since 2011 which reflect the most recent knowledge in the field and get 9,764 articles.

\paragraph{Entity Recognition}
In order to identify knowledge at the proper granularity level for AD, we extract relevant entities from the corpus by utilizing the PubTator Central (PTC)~\cite{wei2013pubtator} developed and continuously maintained by NCBI. PTC is a widely-used tool to provide state-of-the-art annotations of biomedical concepts for PubMed abstracts and full-text articles, and it supports six bioconcept types including genes, diseases, chemicals, mutations, species and cell lines. We apply PTC to the abstracts of all our AD papers and obtain the relevant named entities which will serve as nodes in the knowledge graph.

\begin{figure}[!t]
    \centering
    \includegraphics[width=7.5cm]{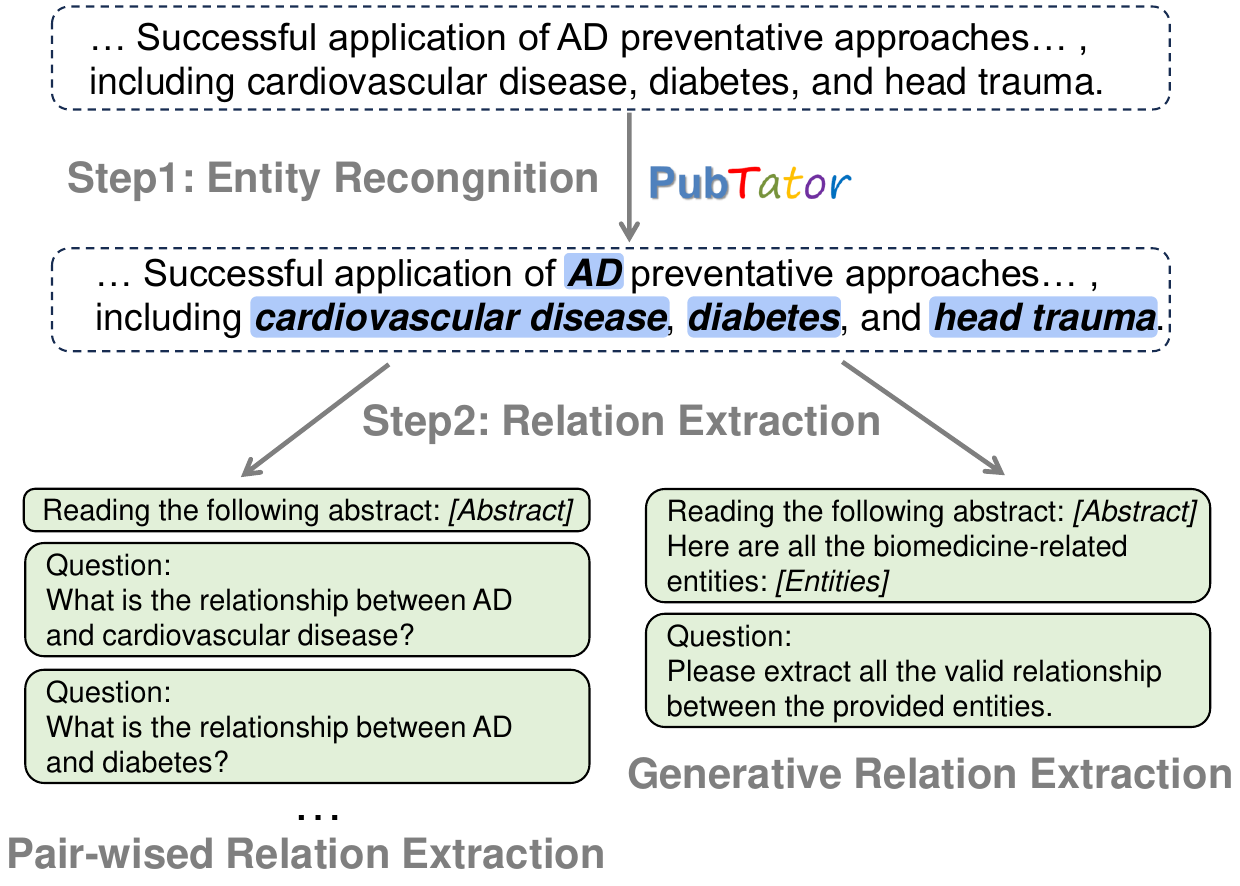}
    \caption{The detailed process of AD-specific KG construction.}
    \label{fig:RE}
\end{figure}

\paragraph{Relation Extraction} To build an accurate and high-quality knowledge graph on AD, we aim to assign a specific relation type between the two related entities. Through a comprehensive survey of relation extraction methods for knowledge graph construction, we categorize current approaches with LLMs into two main groups: (a). \textbf{Pair-wised Relation Extraction}~\cite{carta2023iterative,wadhwa2023revisiting} aims to prompt the LLMs to describe the relationship between any two entities in a segment of text. (b). \textbf{Generative Relation Extraction}~\cite{han2023pive,bi2024codekgc,datta2024construction}, where LLMs directly output all related entity pairs and their corresponding relationships. As shown in Figure~\ref{fig:RE}, we incorporate both of these relation extraction methods into our knowledge graph augmentation process to provide a comprehensive comparison between them. We denote the resulting knowledge graphs from these approaches as $KG_{pair}$ and $KG_{gen}$ respectively.

Table~\ref{statistics} presents the detailed statistics about our augmented knowledge graph, including the number of corpora we used, and the number of nodes, relations and triples in $KG_{pair}$ and $KG_{gen}$.

\subsection{KG for LLMs}
\label{KG for LLMs}

In this section, we begin by outlining our process for sampling coarse-grained augmented knowledge from our evolving knowledge graph (Section~\ref{Coarse-grained Knowledge Sample}). Subsequently, we delve into detail regarding our self-aware knowledge retrieval method, which aims to filter out noise and retrieve the most pertinent knowledge to provide to the LLM (Section~\ref{Self-aware Knowledge Retrieval}).

\subsubsection{Coarse-grained Knowledge Sample}
\label{Coarse-grained Knowledge Sample}

Given a question query $Q$, we first construct a prompt and ask LLMs to extract all the domain-specific entities $E=\{e_1, e_2, ...\}$ from it.
Afterward, we adhere to the methodology proposed by~\citet{wen2023mindmap} and execute a similarity-based entity linking process to connect all entities within $E$ to the entity structure in our knowledge graph $G$. Specifically, we employ a semantic similarity model~\cite{reimers2019sentence} to encode all entities in $G$ and $E$ into dense embeddings, denoted as $H_G$ and $H_E$, respectively. Subsequently, utilizing cosine similarity, we establish links between each entity in $E$ and its nearest neighbor entity in $G$. This procedure yields an initial entity set $E_G$ for the subsequent knowledge sampling step.

To build an evidence sub-graph to boost LLMs' reasoning process, we follow the previous study~\cite{wen2023mindmap} and consider the following two kinds of explorations in our AD-KG:


\paragraph{Path-based Exploration} entails the extraction of a sub-graph from $G$ to encompass all entities within $E_G$. The process unfolds as follows: (a) Begin by selecting one node from $e_Q^0$ as the initial node, denoted as $e_1$, and place the remaining nodes into a candidate node set, $E_{cand}$. Explore at most $k$ hops from $e_1$ to identify the subsequent node, $e_2$, where $e_1 \in E_{cand}$. If $e_2$ is successfully reached within $k$ hops, update the start node to $e_2$ and remove $e_2$ from $E_{cand}$. In the event $e_2$ cannot be found within $k$ hops, concatenate the segment paths acquired thus far and store them in $G_Q^{path}$. Subsequently, choose another node $e_1^\prime$ from $V_{cand}$ as the new start node, and eliminate both the original start node $e_1$ and the current node $e_2$ from $E_{cand}$. (b) Verify if $E_{cand}$ is empty. If not, repeat step (a) to identify the next segment of the path. If $E_{cand}$ is empty, combine all segments to construct a set of sub-graphs and place them into $G_Q^{path}$.

\paragraph{Neighbor-based Exploration} endeavors to augment the evidence relevant to the query within $G_Q$. This process consists of two steps: (a) Initially, expand each node $e$ within $E_G$ by 1-hop to incorporate their neighbors ${e^\prime}$, thus appending triples ${(e, r, e^\prime)}$ to $G_Q^{nei}$. (b) Then assess whether each $e^\prime$ exhibits semantic relevance to the query. If affirmative, further expand the 1-hop neighbors of $e^\prime$, consequently adding triples $(e_{nei}, r^\prime, e^\prime)$ to $G_Q^{nei}$.

After obtaining the two sub-graphs $G_Q^{path}$ and $G_Q^{nei}$, we perform post-processing to further prune redundant information in sub-graphs and prompt LLMs to describe the structure of each sub-graph.

\subsubsection{Self-aware Knowledge Retrieval}
\label{Self-aware Knowledge Retrieval}

In our initial experiment, we noticed the coarse-grained knowledge sampled with the above-mentioned approaches still contained redundant and irrelevant information.
This issue of noise is a common challenge encountered in automatically-constructed knowledge graphs~\cite{fang2021discos,zhang2020aser,li2022c3kg,bi2024codekgc}. Moreover, many recent works~\cite{yu2023chain,li2023compressing,chen2024benchmarking,wu2024easily} have demonstrated LLMs can indeed be influenced by such noisy information.
To address this challenge, we borrow insights from the recent self-powered LLMs~\cite{wang2022self,pan2023automatically,li2023dail,yuan2024self,tong2024can} and LLM-as-a-judge~\cite{Zheng2023JudgingLW}, proposing a self-aware knowledge retrieval method to leverage LLMs' ranking capability~\cite{sun2023chatgpt,ma2023large} to filter out noisy information.

In particular, we directly prompt the LLM to rerank the sampled knowledge and only retrieve top $k$ triples to provide for itself in the final-round inference. Given the question $Q$ and either the path-based or neighbor-based sub-graph $G_Q$, we create prompt $p_{self}$ by filling the pre-defined template:
\begin{equation}
    p_{self} = {\rm Template_{self}}(Q, G_Q, k).
\end{equation}
Then, we use $p_{self}$ as the input to prompt the LLM to obtain the self-retrieved knowledge:
\begin{equation}
   G_Q^{self} = {\rm LLM}(p_{self}),
\end{equation}
Finally, we provide the question $Q$ and fine-grained knowledge $G_Q^{self}$ to the LLM for reasoning and get the predicted answer $a$ in two steps:
\begin{equation}
    p_{inference} = {\rm Template_{inference}}(Q, G_Q^{self}),
\end{equation}
\begin{equation}
   a = {\rm LLM}(p_{inference}).
\end{equation}
We provide detailed examples in Appendix~\ref{Details of LLMs for KG} and \ref{Details of KG for LLMs} to demonstrate the input and output in our DALK.

%% file: latex/4-main-experiment.tex
\subsection{ADQA Benchmark}
For performance evaluation, we consider four widely-used medical QA datasets spanning diverse biomedical topics~\cite{jin2021disease, ankit2022medmcqa, hendrycks2021measuring, anselmo2013qa4mre} and derive an AD-specific QA dataset from them. The four medical QA datasets are all multiple-choice based and include: 1) MedQA~\cite{jin2021disease} consisting of US Medical Licensing Examination (USMLE)-style questions, 2) MedMCQA~\cite{ankit2022medmcqa} containing medical school entrance exam questions from India, 3) MMLU~\cite{hendrycks2021measuring} consisting of diverse biomedical and clinical questions from various sources, 4) QA4MRE~\cite{anselmo2013qa4mre} containing a subset of questions for AD derived from PubMed and Medline. In order to extract from the medical QA datasets a subset of samples related to AD for our evaluation, we referred to NIH's Common Alzheimer's and Related Dementias Research Ontology (CADRO) \footnote{https://iadrp.nia.nih.gov/about/cadro}. Jointly developed by the National Institute on Aging and the Alzheimer’s Association, CADRO is a three-tiered classification system with eight main categories and a dozen sub-categories for AD and related dementia, and it contains common terminologies or keywords used in the field. We derived from the CADRO a list of AD-related keywords most relevant to the medical QA datasets: <Aging, Alzheimer, Amyloid beta, APOE, Dementia, Lipoprotein, Microglia>. Then, we searched against each medical QA dataset for matches with these keywords to find putative QA samples, then further asked GPT-3.5-turbo to judge for each putative sample whether the question is indeed related to AD or not. Finally, we filtered out a subset of such samples that are considered highly relevant to AD to conduct our evaluation (number of samples in each dataset is shown in Table~\ref{Main result-tab}).
More details can be found in Appendix~\ref{ADQA Benchmark}.

\subsection{Experiment Settings}
We apply our framework with OpenAI GPT-3.5-turbo models~\cite{Introducing}. We also include the following baseline methods for comparison:

\paragraph{Biomedical LLMs}
Both ChatDoctor~\cite{yunxiang2023chatdoctor} and Med-Alpaca~\cite{shu2023visual} are fine-tuned versions of LLaMA~\cite{touvronllama} on biomedical corpora.
Compared with them, Meditron~\cite{chen2023meditron} is built on LLaMA-2~\cite{touvron2023llama} and extends its pretraining on a comprehensively curated medical corpus.
BiomedGPT~\cite{zhang2023biomedgpt} is also based on LLaMA-2 and pioneer as the first open-source and generalist visual language AI for diverse biomedical tasks.
Biomistral~\cite{labrak2024biomistral} is an LLM crafted specifically for the biomedical domain, optimized for efficiency through quantization and model merging techniques.

\begin{table*}[h]\small
\centering
\begin{tabular}{lccccc}
\toprule[1.2pt]                                              & MedQA      & MedMCQA      & MMLU       & QA4MRE     & AVG           \\ \hline
\multicolumn{6}{l}{\emph{\textbf{Biomedical LLMs}}}                                                                                                                        \\
ChatDoctor-7B~\cite{yunxiang2023chatdoctor}                                                                 & 25.7          & 36.4          & 46.9          & 51.4          & 40.1          \\
Med-Alpaca-7B~\cite{shu2023visual}                                                                  & 41.4          & 42.8          & 44.9          & 57.1          & 46.5          \\
BiomedGPT-7B~\cite{zhang2023biomedgpt}                                                                   & 38.8          & 41.9          & 48.9          & 42.6          & 43.1          \\
Meditron-7B~\cite{chen2023meditron}                                                                    & 27.6          & 31.4          & 36.7          & 25.7          & 30.4          \\
Biomistral-7B~\cite{labrak2024biomistral}                                                                  & 44.7          & 49.5          & 53.1          & 68.6          & 54.0          \\
Meditron-70B                                                                 & 50.0          & 44.8          & 79.6        & 51.4          & 56.4          \\
ClinicalCamel-70B~\cite{toma2023clinical}                                                                  & 50.0          & 64.3          & \underline{83.7}          & 68.6          & 66.7          \\ \hline
\multicolumn{6}{l}{\emph{\textbf{Retrieval-augmented LLMs}}}                                                                                                                                    \\
GPT-3.5-turbo w/ Ada~\cite{NewEmbedding}                                                                 & \underline{57.2}          & 65.7          & \underline{83.7}          & 62.9          & 67.4          \\
Almanac~\cite{zakka2024almanac}                                                             & 48.0          & 69.5          & 71.4          & 60.0          & 62.2          \\
Clinfo.ai~\cite{lozano2023clinfo}                                                           & 54.3          & \textbf{77.0} & 81.3          & 67.7          & \underline{70.1}          \\
\begin{tabular}[c]{@{}l@{}}Clinfo.ai w/o PubMed API\end{tabular} & 49.3          & 68.6          & 79.6          & \textbf{74.3} & 67.9          \\ \hline
GPT-3.5-turbo                                                                    & 50.0          & 71.9          & 83.6          & 62.9          & 67.1          \\
\textbf{DALK}                                                               & \textbf{57.9} & \underline{75.2}          & \textbf{85.4} & \underline{71.4}          & \textbf{72.6} \\
\toprule[1.2pt]
\end{tabular}
\caption{Experiment results on our constructed ADQA benchmark. The best results of each metric are in bold and the second-best results are underlined. The ``AVG'' column represents the average accuracy score on the four sub-dataset.}
\label{Main result-tab}
\end{table*}

\paragraph{Retrieval-Augmented LLMs}
Furthermore, we also compare our method with several representative retrieval-augmented LLMs in the biomedical domain. Almanac~\cite{zakka2024almanac} is a novel approach utilizing OpenAI's GPT model integrated with a Qdrant vector database to hold external sources of knowledge retrieved from local corpus, web search, and calculators, designed to answer open-domain clinical questions. Like Almanac, \citet{lozano2023clinfo} introduced Clinfo.ai, which is an open-source, end-to-end retrieval-augmented LLM (GPT) to answer medical queries using scientific literature summarizations derived from PubMed search engine. We adopt both Almanac and Clinfo.ai with the same prompt as ours to answer multiple-choice questions to suit the ADQA benchmark. Lastly, we implement a simple retrieval-augmented GPT baseline with CoT prompting similar to our proposed DALK. All the GPT models used are set to GPT-3.5-turbo as detailed in the next paragraph, to be consistent.

\paragraph{Implementation Details}
We use the knowledge graph constructed with the generative approach ($KG_{gen}$) in our main experiment and conduct an ablation study on the knowledge graph with RE method ($KG_{pair}$) in Section~\ref{Ablation Study on KG Construction}.
We use GPT-3.5-turbo with the version ``gpt-3.5-turbo-0301'' and set the sampling temperature to 0.7.
We utilize 7B versions of all the biomedical LLMs baselines.
For RAG methods, we split each document with a max length of 128 and retrieve the top 3 most relevant documents as the support evidence for LLMs to do inference.
We set the parameter $k$ in our self-aware knowledge retrieval to 5 and conduct further analysis on it in Section~\ref{Hyper-parameter Analysis}.

\subsection{Main Result}


Table~\ref{Main result-tab} shows the experimental results on our ADQA benchmark. We note that upon applying our dynamic co-augmentation framework, DALK's performance surpasses that of other biomedical LLMs and RAG methods overall. It consistently achieves either the best or the second-best accuracy score across all sub-datasets and attains the highest AVG score. Furthermore, the substantial improvement over vanilla GPT-3.5-turbo underscores the efficacy of our approach in domain-specific ADQA.

Furthermore, we observe that the performance of biomedical-specific LLMs generally lags behind that of GPT-3.5-turbo. We attribute this discrepancy to the smaller size of these biomedical LLMs. While they may perform adequately in general medical contexts, they fall short in the AD scenario, which demands more domain-specific knowledge. In the case of GPT-3.5-turbo combined with various RAG methods, it is evident that most RAG methods enhance the models' performance. Among them, GPT-3.5-turbo with Clinfo.ai yields the most significant improvement, boosting the accuracy score from 67.1 to 70.1 compared to vanilla GPT-3.5-turbo. However, it is important to note that the original Clinfo.ai necessitates access to the PubMed API, constituting an external resource. When we disable this access and solely utilize the same corpora as in DALK within the Clinfo.ai retrieval system, the improvement it brings becomes marginal and incomparable to our method.
Due to the space limitation, we put more RAG results with different hyper-parameters in Appendix~\ref{Further Experiment for RAG}.

\subsection{Ablation Study on Self-aware Knowledge Retrieval}
\label{Ablation Study on Self-aware Knowledge Retrieval}

\begin{table}[h]
\tiny
\centering
\setlength{\tabcolsep}{1.8mm}{
\begin{tabular}{lccccc}
\toprule[1.2pt]
& MedQA         & MedMCQA       & MMLU          & QA4MRE        & AVG           \\ \hline
AVG Length     & 107.4          & 23.8          & 342.9          & 17.6          & 122.9          \\ \hline
GPT-3.5-turbo                                                                                    & 50.0          & 71.9          & 83.6          & 62.9          & 67.1          \\
\textbf{DALK}                                                                                & \textbf{57.9} & \textbf{75.2} & \textbf{85.4} & 71.4          & \textbf{72.6} \\
\begin{tabular}[c]{@{}l@{}}DALK\\\: w/o self-aware\\\: knowledge retrieval\end{tabular} & 56.5          & 71.0          & 77.6          & \textbf{77.1} & 70.6 \\ \toprule[1.2pt]
\end{tabular}}
\caption{Ablation study results with and without our proposed self-aware knowledge retrieval.}
\label{ablation:self}
\end{table}


In this section, we evaluate the efficacy of our proposed self-aware knowledge retrieval method through an ablation study. As depicted in Table~\ref{ablation:self}, we observe that while the dynamic co-augmentation framework without the self-aware knowledge retrieval module still enhances the model's performance, the overall improvement is less pronounced.
Furthermore, we observe that the efficacy of self-aware knowledge retrieval correlates with the length of queries within a given context. For instance, a notable enhancement in performance is evident within the MMLU sub-dataset upon the implementation of self-aware knowledge retrieval.
We attribute this to the fact that questions in the MMLU dataset typically contain longer contexts compared to other medical QA datasets integrated into ADQA. Consequently, irrelevant knowledge sourced from the context may exacerbate the issue of information noise thus underscoring the necessity for self-aware retrieval.
Conversely, within QA4MRE, characterized by shorter query lengths, the application of self-aware knowledge retrieval can even lead to a decline in performance.

\subsection{Ablation Study on KG Construction}
\label{Ablation Study on KG Construction}



\begin{table}[h!]
\small
\centering
\begin{tabular}{lcc}
\toprule[1.2pt]
& AVG           & \#Triples \\ \hline
GPT-3.5-turbo                                                                  & 67.1          & -         \\ \hline
\begin{tabular}[c]{@{}l@{}}\textbf{DALK}\\  w/ Generative KG\end{tabular} & \textbf{72.6} & 53,585    \\
\begin{tabular}[c]{@{}l@{}}\textbf{DALK}\\  w/ RE KG\end{tabular}         & 66.3          & 171,431  \\ \toprule[1.2pt]
\end{tabular}
\caption{Ablation study results with generative construction and RE construction.}
\label{ablation: kg}
\end{table}

Table~\ref{ablation: kg} illustrates the results of the ablation study conducted using generatively constructed KG and RE-constructed KG. Surprisingly, despite the RE method yielding a KG with a larger scale and more triples, knowledge sampled from it has unexpectedly resulted in a non-trivial drop in performance within ADQA.
After a manual examination of the two constructed knowledge graphs, we find LLMs with the RE construction method have a strong inclination to wrongly assign a relationship to two unrelated entities, which has been exposed by the previous studies~\cite{wan2023gpt}.
In contrast, the generative construction approach exclusively outputs triples that LLMs confidently endorse, yielding a smaller yet more precise knowledge graph. This trade-off between coverage and accuracy underscores the critical importance of denoising in the construction of KGs by LLMs.

%% file: latex/5-further-analysis.tex
\subsection{Co-augmentation Analysis}

\begin{figure}[!h]
    \includegraphics[width=8cm]{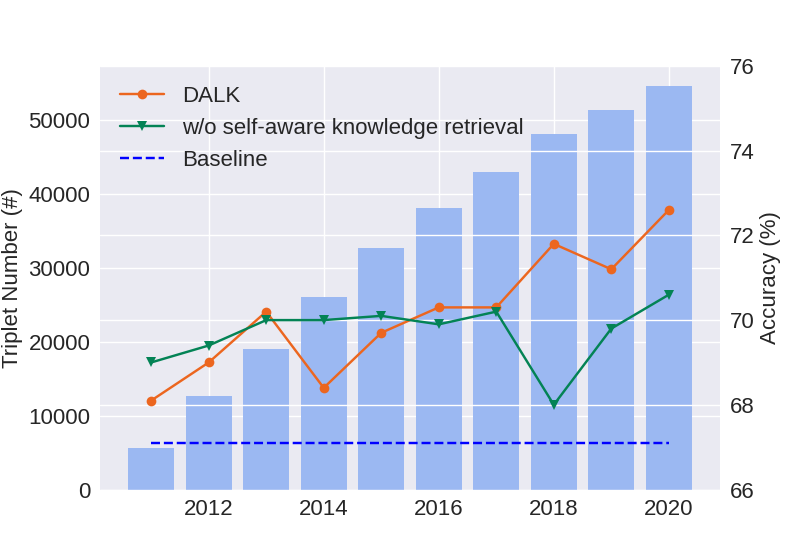}
    \caption{The size of the knowledge graph (triplet number) and the KG-augmented GPT-3.5-turbo's performance (accuracy) over time.}
    \label{evolution}
\end{figure}

To comprehensively understand how the performance of LLMs evolves in response to increasing KG sizes, we undertake a detailed co-augmentation analysis. Illustrated in Figure~\ref{evolution}, our experiments aim to discern the changing performance trends of LLMs as the knowledge triples accumulate annually. Our findings reveal that our framework effectively fosters the co-evolution of LLMs and KG, with the performance of KG-augmented LLMs exhibiting a generally upward trajectory as the KG expands.
Notably, when we remove the self-aware knowledge retrieval module, this upward trend becomes less significant.
This further implies the importance of sampling and selecting appropriate knowledge for LLMs when the KG's size increases.

\begin{table*}[h!]
\centering
\small
\begin{tabular}{llc}
\toprule[1.2pt]
& Path-based Sub-graph                                                                                                                                                                                                                                                                                                                                                                & Answer \\ \hline
Baseline                                                                                 & -                                                                                                                                                                                                                                                                                                                                                                                   & C \XSolidBrush     \\ \hline
\begin{tabular}[c]{@{}l@{}}DALK\\   -w/o self-aware\\   knowledge retrieval\end{tabular} & \begin{tabular}[c]{@{}l@{}}neurofibrillary tangles-\textgreater{}FORM BY-\textgreater{}microtubule-associated protein tau...\\ ...\\ entorhinal cortex-\textgreater{}is a part of-\textgreater{}brain-\textgreater{}ASSOCIATES-\textgreater{}mouse with\\ Alzheimer's disease-\textgreater{}brain region-\textgreater{}temporal lobe\\ \end{tabular} & C \XSolidBrush     \\ \hline
\textbf{DALK}                                                                                     & \begin{tabular}[c]{@{}l@{}}Reranked Triples1: entorhinal cortex -\textgreater is a part of -\textgreater brain\\ Reranked Triples2: entorhinal cortex -\textgreater associates -\textgreater mouse with Alzheimer's disease\\ Reranked Triples3: temporal lobe -\textgreater affected by -\textgreater Alzheimer's disease\end{tabular}                                             & D \Checkmark  \\ \toprule[1.2pt]  
\end{tabular}
\caption{An example for the case study. The question is: ``The area of the brain resistant to Neurofibrillary tangles of Alzheimer's disease is: A. Visual association areas B. Entorhinal coex C. Temporal lobe D.Lateral geniculate body''}
\label{tab:case study}
\end{table*}

\subsection{Hyper-parameter Analysis}
\label{Hyper-parameter Analysis}

\begin{figure}[!h]
    \includegraphics[width=8.3cm]{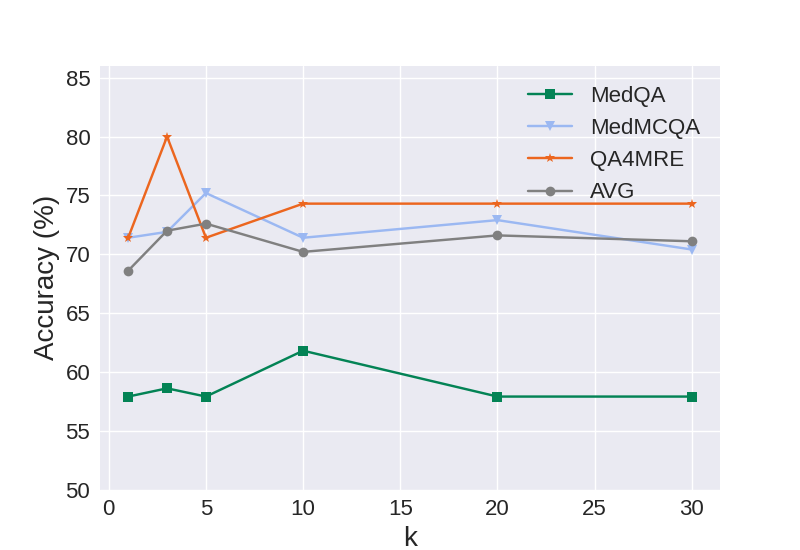}
    \caption{Different self-aware knowledge retrieval settings in MedQA, MedMCQA, QA4MRE and the average score (AVG).}
    \label{hyper-parameter}
\end{figure}

In this section, we do a hyper-parameter analysis on the retrieval number $k$ of our self-aware retrieval module.
We select a group of value for $k$ ([1,3,5,10,20,30]) and present the experiment results in Figure~\ref{hyper-parameter}.
We show the accuracy score on MedQA, MedMCQA, QA4MRE and AVG with different $k$.
We find when $k$ is small, an increment to it can lead to a performance enhancement.
After the best performance shows up, continually increasing the value of $k$ will cause a smooth decrease in the model accuracy score.
This result indicates the knowledge ranked in the top positions is more helpful while the knowledge ranked behind is not very useful, thus successfully validating the capability of LLMs to do fine-grained knowledge reranking.

Moreover, we find the best $k$ value is correlated with the averaged query length.
For example, the best performance in MedQA (average query length is 107.4) appears when $k=10$ while the best performance in MedMCQA and QA4MRE appears when $k=5$ and $3$ respectively.
This aligns with our previous finding that a longer query corresponds to a larger and noisier sub-knowledge graph.

\subsection{Sensitivity Analysis on ADQA Benchmark}

\begin{table}[!h]
\centering
\small
\begin{tabular}{lcc}
\toprule[1.2pt]
Benchmark        & \textbf{DALK} & \begin{tabular}[c]{@{}l@{}}\textbf{DALK}\\\: w/o self-aware\\\: knowledge retrieval\end{tabular}           \\ \hline
ADQA                 &  72.6 & 70.6          \\
w/o ``Alzheimer''    &  72.1 & 70.4 \\
w/o ``Dementia''     &  72.4 & 71.3          \\
w/o ``APOE''         &  73.2 & 71.2          \\
w/o ``Amyloid beta'' &  73.5 & 70.7          \\
w/o ``Aging''        &  72.9 & 71.4          \\
w/o ``Lipoprotein''  &  73.1 & 71.0          \\
w/o ``Microglia''    &  72.8 & 70.9          \\ \toprule[1.2pt]
\end{tabular}
\caption{Sensitivity analysis for ADQA benchmark with a leave-one-out evaluation on AD-related keywords.}
\label{tab:sensitivity analysis}
\end{table}

In this section, we conduct a sensitivity analysis for our constructed ADQA by conducting a leave-one-out evaluation on AD-related keywords.
We do it by removing the samples with each keyword in our keyword list and calculating the AVG score of the remaining samples.
As the result shown in Table~\ref{tab:sensitivity analysis}, we find not all of the keywords are incorporated in our ADQA benchmark.
Notably, the keywords ``CSF Biomarkers'', ``Neurogenesis'', ``PET Amyloid'', ``PET Tau'', ``Tau Phosphorylation'' lack corresponding samples in ADQA.
We believe one critical work in the future for benchmarking AD-related knowledge is to collect QA samples to cover these missing keywords.
Moreover, analyzing the performance variation upon removing samples linked to each keyword offers insight into determining the relevance of the keyword to AD.

\subsection{Case Study}
\label{Case Study}
We put an example in Table~\ref{tab:case study} to showcase the efficacy of DALK.
We notice while the path-based sub-graph contains the relevant knowledge to exclude option C, it still involves other irrelevant information and finally fails to prompt the LLMs to produce the correct answer.
In contrast, our self-aware knowledge retrieval method successfully chooses the top 3 most relevant triples for the given problem and results in the correct answer D.

\subsection{Experiment Results with GPT-4}

\begin{table}[]\centering
\setlength{\tabcolsep}{1.2mm}{
\begin{tabular}{lcc}
\toprule[1.2pt]
                    & MedMCQA & MMLU \\ \hline
GPT-4-turbo         & 88.1    & 81.3 \\
GPT-4-turbo w/ DALK & \bf{89.2}    & \bf{91.8} \\ \toprule[1.2pt]
\end{tabular}}
\caption{Experiment Result in GPT-4.}
\label{exp:gpt4}
\end{table}

To evaluate our method in the SOTA LLM, we select two subsets, MedMCQA and MMLU and conduct the experiment with GPT-4-turbo. As the result shown in the Table~\ref{exp:gpt4}, we find in both subsets, DALK achieves significant performance enhancement. Another thing to note is in the MMLU dataset, vanilla GPT-4-turbo (81.3) is even worse than GPT-3.5-turbo (83.6). We will conduct more experiments in larger models in the future to thoroughly explore our proposed method.

\subsection{Generalization of DALK}

\begin{table}[]\centering
\begin{tabular}{lc}
\toprule[1.2pt]
                      & GPT-4 Score \\ \hline
GPT-3.5-turbo         & 0.61             \\
GPT-3.5-turbo w/ DALK & \bf{0.66}             \\ \toprule[1.2pt]
\end{tabular}
\caption{Experiment Result in COVID-QA.}
\label{exp:covid}
\end{table}

To demonstrate the robustness and generalization of our DALK, We conduct a further evaluation with COVID-QA, an open-ended QA dataset focused on COVID-19. For efficiency, We use a small set of literature abstracts from the CORD-19 dataset as the raw corpus to build KG. We conduct the experiment using GPT-3.5-turbo from OpenAI. For evaluation, we follow some previous work and prompt GPT-4 to score each generated answer. As the result shown in Table~\ref{exp:covid}, DALK also significantly improves model performance in this open-ended QA dataset focuses on COVID.

%% file: latex/6-conclusion.tex
In this research, we begin by analyzing the main limitations of adopting the existing LLMs-based methods in AD-specific areas.
To address these issues, we propose a novel approach in the merging of large language models and knowledge graphs in the context of Alzheimer's Disease. We provide an innovative dynamic co-augmentation framework for the refinement of large language models and knowledge graphs.
Initially, our approach extracts structural insights from the unstructured scientific literature, crafting a specialized knowledge graph for AD. Subsequently, we employ a coarse-to-fine sampling technique coupled with a unique self-aware knowledge retrieval strategy to pinpoint relevant information from the knowledge graph.
The extensive evaluation conducted in our constructed ADQA benchmark shows the effectiveness of our method and provides further hints into the synergy of LLMs and KG in the context of AD.

%% file: latex/7-limitations.tex

In the development of our AD-KG, our primary focus lies in the exploration of two distinct methods for extracting relationships between associated entities. For entity recognition, we employ a strong PubTator annotator directly, without delving into the utilization of LLMs in this context. However, we have observed that LLMs also exhibit promising entity extraction capabilities in Section~\ref{Coarse-grained Knowledge Sample}. We defer the refinement of methods for extracting entities for KG construction with LLMs to future works. 
Furthermore, a significant contribution of our work is the establishment of the ADQA benchmark. Nonetheless, the datasets utilized in constructing ADQA primarily consist of medical school exam questions, potentially exhibiting a domain gap from the scientific literature informing AD-KG. One potential remedy is leveraging PubmedQA~\cite{jin2019pubmedqa}; however, it is hindered by limited data amount. In the future, we will keep gathering AD-related QA samples and expanding the size of our ADQA benchmark.
In the future, we will do more exploration in adopting and benchmarking LLMs in the AD areas.

%% file: latex/8-ethics-statement.tex
We have familiarized ourselves with and honour the ethical code set out in the ACL Code of Ethics\footnote{https://www.aclweb.org/portal/content/acl-code-ethics}.
The knowledge graphs constructed in the paper are based on published scientific literature from PubMed. 
The ADQA dataset used in the study is also derived from publicly available medical QA datasets that are properly cited. 
We strive to ensure our study upholds ethical principles and not cause any kind of safety or privacy concerns.
Although not observed in our multiple-choice QA analysis, we recognize the possibility of factual errors and hallucinations when using pre-trained LLMs for medical QA tasks in general, and we do not recommend these models be applied in a practical setting at present.